\pdfoutput=1 
\documentclass[10pt,twocolumn,letterpaper]{article}

\usepackage{cvpr}              
\usepackage{graphicx}
\usepackage{amsmath}
\usepackage{amssymb}
\usepackage{booktabs}
\usepackage{verbatim}
\usepackage{url}
\usepackage{hyperref}

\usepackage[capitalize]{cleveref}
\crefname{section}{Sec.}{Secs.}
\Crefname{section}{Section}{Sections}
\Crefname{table}{Table}{Tables}
\crefname{table}{Tab.}{Tabs.}

\def\cvprPaperID{12000}
\def\confName{CVPR}
\def\confYear{2023}

\begin{document}
\title{Bayesian posterior approximation with stochastic ensembles}

\author{Oleksandr Balabanov$^{1}$,  Bernhard Mehlig$^{2}$,  Hampus Linander$^{2,3}$\\ 
$^{1}$Stockholm University\hspace{0.5cm} $^{2}$University of Gothenburg\hspace{0.5cm}
$^{3}$Chalmers University of Technology\\
{\tt\small oleksandr.balabanov@fysik.su.se, bernhard.mehlig@physics.gu.se,
hampus.linander@gu.se}\\
{\tt\small Code: github.com/oleksandr-balabanov/stochastic-ensembles}
}

\maketitle

\begin{abstract}

We introduce ensembles of stochastic neural networks to approximate the Bayesian posterior, combining stochastic methods such as dropout with deep ensembles. The stochastic ensembles are formulated as families of distributions and trained to approximate the Bayesian posterior with variational inference. We implement stochastic ensembles based on Monte Carlo dropout, DropConnect and a novel non-parametric version of dropout and evaluate them on a toy problem and CIFAR image classification. For both tasks, we test the quality of the posteriors directly against Hamiltonian Monte Carlo simulations. Our results show that stochastic ensembles provide more accurate posterior estimates than other popular baselines for Bayesian inference.

\end{abstract}

\section{Introduction}
\label{sec:intro}
Bayesian neural networks provide a principled way of reasoning about model selection and assessment with posterior distributions of model parameters \cite{Denker1987, Tishby1989, Buntine1991, MacKay1991}. Although the analytical Bayesian posteriors can answer questions about model parameter uncertainty, the immense computational challenge for commonly used neural network architectures makes them practically infeasible\footnote{There are examples of closed-form solutions for small \text{architectures~\cite{wu2018deterministic}.}}. Instead, we are forced to resign to approximation methods that makes a trade-off between posterior accuracy and computational complexity \cite{bnn_review}. 

A prominent method to approximate the Bayesian posterior is deep ensembles~\cite{Wilson2020, Hansen1990, Lakshminarayanan2016}, that can be shown to correspond to a variational inference approximation with a delta distribution family \cite{Hoffmann2021}. This method is implemented by simply training an ensemble of models and treating them as samples from the model posterior. In the  variational inference formulation, this corresponds to approximating the posterior by sampling from many sets of maximum a posteriori parameters. 

To further reduce the computational effort in evaluating the approximate posterior, stochastic methods such as Monte Carlo dropout \cite{Srivastava2014} and DropConnect \cite{Wan2013} inference have also been used extensively \cite{Gal2015, Gal2016, Mobiny2019}. They benefit from computationally cheaper inference by virtue of sampling stochastically from a single model. Formulated as a variational approximation to the posterior, dropout samples from a family of parameter distributions where parameters can be randomly set to zero. Although this particular family of distributions might seem unnatural \cite{Folgoc2021}, it turns out that the stochastic property can help to find more robust regions of the parameter space, a fact well-known from a long history of using dropout as a regularization method.

Recently, there has been great progress towards understanding the analytical posteriors of larger neural networks by means of direct Markov Chain Monte Carlo sampling of the parameter posterior \cite{Izmailov2021}. Through impressive computational efforts, the posteriors for models as large as ResNet-20 have been sampled using Hamiltonian Monte Carlo (HMC) simulations. The sampled posterior has been shown to provide more accurate predictions that are surprisingly sensitive to data distribution shift as compared to standard maximum likelihood estimation training procedures. These HMC computations have made it possible to compare approximation methods such as dropout inference and deep ensembles directly to the Bayesian posterior. Ensembling and stochastic methods such as dropout have been used successfully to find posterior approximations in many settings, but without a Bayesian formulation that includes both ensembling and stochastic methods it is difficult to understand if and how the two approaches can complement each other.

Recent work have shown that uncertainty quantification is subjective to neural network architectures, and that the accuracy of posterior approximations depend non-trivially on model architecture and dataset complexity \cite{HLOBHYBM2022}. To find computationally efficient methods that can accurately approximate the Bayesian posterior, for different data domains and architectures, is therefore an important goal with practical implications for applications that require accurate uncertainty quantification to assess network predictions\cite{Gawlikowski2022survey, Moloud2021}.

In this paper we combine deep ensembles and \text{stochastic} regularization methods into stochastic ensembles of neural networks. We formulate the stochastic ensemble construction within the Bayesian variational formalism, where multiple stochastic methods such as regular Monte Carlo dropout, DropConnect, and others are combined with ensembling into one general variational ansatz. We then conduct a series of tests using a simple toy model (synthetic data) and CIFAR (image classification), where stochastic deep ensembles are found to provide more accurate posteriors than MultiSWA~\cite{Wilson2020} and regular deep ensembles in a number of settings. In particular, for CIFAR we use a neural network architecture evaluated by Izmailov et al. \cite{Izmailov2021} in their large-scale experiments, allowing us to make a direct comparison of the ensemble methods to the ``ground truth" HMC posterior.

\section{Related Work}

 Bayesian inference for neural networks has a long history \cite{Denker1987, Tishby1989, Buntine1991, MacKay1991}. A variety of methods have been developed for estimating the Bayesian posterior, including Laplace approximation\cite{MacKay1992book}, Markov Chain Monte Carlo methods \cite{Neal1996, Welling2011, Springenberg2016}, variational Bayesian methods~\cite{Hinton1993, Barber98, Graves2011, Blundell2015, Kingma2015, Louizos2017, Wu2018} and Monte Carlo dropout~\cite{Gal2015, Gal2016}. Deep ensembles~\cite{Hansen1990, Lakshminarayanan2016} have shown surprising abilities to approximate the Bayesian posterior well. They were found to empirically outperform other scalable methods~\cite{Gustafsson2019, Ovadia2019} and important conceptual aspects of their excellent performance were recently explained  \cite{Lakshminarayanan2019, Wilson2020, Dwaracherla2022}. 
Forced diversity between ensemble members, motivated from a Bayesian perspective, has been shown to improve accuracy and out-of-distribution detection \cite{Angelo2021}. The dependence of the posterior on priors in weight and function space for deep ensembles has also been investigated~\cite{tiulpin2021greedy,Angelo2021}.
 Various methods for improving deep ensembles have also recently been proposed~\cite{Stickland2020, Nam2021, Mehrtens2022, Zaidi2021, Kook2022, Bui2020, Wenzel2020, Wen2020}.

 One of the widely used ensemble methods is MultiSWAG~\cite{Wilson2020}, where an ensemble of networks trained with Stochastic Weight Averaging 
Gaussian~\cite{Izmailov2018, Maddox2019, Wilson2020} is used for approximating the posterior. These networks were shown to have better generalization properties than networks trained using  conventional SGD protocols~\cite{Izmailov2018, Maddox2019}. MultiSWAG was also found to provide posteriors that are more robust to distributional shifts than regular deep ensembles~\cite{Wilson2020}.

Our variational inference formulation is based on the works of Gal et al.~\cite{Gal2015, Gal2016} for dropout and  Hoffmann et al.~\cite{Hoffmann2021} for deep ensembles. We combine these two methods and formulate one general Bayesian variational ansatz. We also note that our independent rederivation of the loss from Ref.~\cite{Hoffmann2021} uncovered a new loss contribution promoting member diversity. A similar loss term was studied in Ref.~\cite{Angelo2021} that was argued to be seen as Wasserstein gradient descent on the Kullback-Leibler divergence. Here we derive it explicitly from Bayesian variational inference.

The numeric tests on CIFAR image classification follow closely the impressive effort of Izmailov at al.~\cite{Izmailov2021} reporting full-batch HMC computations on CIFAR datasets. We reproduced the exact network architecture and used their reported results for making a quantitative comparison of different ensemble types against HMC, indicating that our stochastic ensembles produce better posterior approximations than the non-stochastic baselines.

\section{Our contribution}

\begin{itemize}

\item We introduce stochastic ensembles. They are formulated using Bayesian variational inference  where deep ensembles and stochastic methods such as dropout are combined into one general variational family of distributions. We show the theoretical advantage in using stochastic over regular (non-stochastic) ensembles and test different ensemble methods on a toy problem and CIFAR image classification.

\item We show that for image classification on CIFAR, our stochastic ensembles based on Monte Carlo dropout produce the most accurate posterior approximations compared to other ensemble methods: They are found to be closer to HMC than MultiSWA and regular deep ensembles in terms of log-likelihood loss, accuracy, calibration, agreement, variance, out-of-distribution detection, predictive entropy, and robustness to distributional shifts.

\item The posteriors produced by the Monte Carlo dropout ensembles for CIFAR image classification are also seen to outperform various stochastic gradient Monte Carlo (SGMCMC) methods: They are closer to the HMC posteriors and require less inference samples at test time.

\item We introduce a novel stochastic model with a non-parametric version of dropout by jointly training an ensemble of models with exchange of parameters. It was found to outperform other methods for our simple toy task in terms of the agreement and variance computed in respect to the HMC posterior. It was also found to provide more accurate uncertainty estimates.

\end{itemize}

\section{Bayesian deep learning}

In modelling we aim to predict a label $y^*$ for a given input $x^*$ and provide uncertainty estimates of the prediction. The output should be in alignment with our observation, i.e. it should be conditioned on the training (observed) data $\mathcal{D} = \{x_i, y_i\}^N_{i=1}$. This is captured by the predictive probability distribution $p(y^*| x^*, \mathcal{D})$, which provides a distribution over possible output values $y^*$. 

Consider a model with \text{parameters}~$\theta$. The predictive distribution $p(y^*| x^*, \mathcal{D})$ can then be represented as follows 
\begin{align} 
\begin{split} 
p(y^*| x^*, \mathcal{D})   &= \int \, d\theta \, \underbrace{p(y^*| x^*, \theta)}_\text{aleatoric} \, \underbrace{p(\theta| \mathcal{D})}_\text{epistemic}. \\
\end{split}
\label{eq:Baysian}
\end{align}
The posterior distribution $p(\theta| \mathcal{D})$ describes \text{epistemic} uncertainty, the type of uncertainty that is by definition associated with the model parameters $\theta$. The term $p(y^*| x^*, \theta)$ corresponds to aleatoric (data) uncertainty~\cite{review2022}. 
\\

\textbf{Approximate methods.} In practice, $p(\theta| \mathcal{D})$ is unknown and hard to \text{compute}: \text{Direct} methods for computing $p(\theta| \mathcal{D})$ exist but they are \text{intractable} for models with large number of \text{parameters}. Approximate Bayesian inference aims to find a distribution $q(\theta)$ that is as close to the true posterior $p(\theta| \mathcal{D})$ as possible. To date, the golden standard method for approximate Bayesian inference is HMC\cite{Neal1996}. It is based on performing Metropolis-Hastings updates using Hamiltonian dynamics associated with the potential energy $U(\theta) = - \log p(y |x, \theta)p(\theta)$ with $\mathcal{D} = (x, y)$. This method is accurate but numerically tractable only for small models and datasets, severely limiting its practical applicability.

Variational inference~\cite{Barber1998, Graves2011, Blundell2015} is less precise than HMC, but more numerically efficient. The idea is to look at a family of distributions $q_\omega(\theta)$, where $\omega$ parameterizes the family, and find a distribution that approximates the posterior $p(\theta| \mathcal{D})$ well. One common approach  for finding the best family member $q_\omega(\theta)$ is to numerically minimize the Kullback-Leibler (KL) divergence  $\text{KL}(q_\omega(\theta)\,||\,p(\theta|\mathcal{D}) )$ \cite{review2022}, a measure quantifying how much information is lost when $q_\omega(\theta)$ is approximated by $p(\theta|\mathcal{D})$:
\begin{align} 
\begin{split} 
&\text{KL}(q_\omega(\theta)\,||\,p(\theta|\mathcal{D}) ) =\int \, d\theta \,  q_\omega(\theta) \, \log\,\frac{q_\omega(\theta)}{p(\theta|\mathcal{D})}  \\
&= \text{KL}(q_\omega(\theta)\,||\,p(\theta) ) - \mathbb{E}_{q_\omega(\theta)} [\log \, p(y | \theta, x)] + C,\\
\end{split}
\label{eq:VI}
\end{align}
where the training data is denoted by $\mathcal{D} = (x, y)$ and $C$ is a $\theta$-independent term. Here we applied Bayes' theorem $p(\theta|x,y)  = p(y |\theta, x) \, p(\theta) / p(y |x)$. By minimizing the loss function in Eq.~(\ref{eq:VI}) one obtains an approximation for the posterior $p(\theta|\mathcal{D})$. Note that the cost function consists of two terms, (1) the KL divergence against the prior $p(\theta)$ and (2) the expected negative log likelihood (ENLL), that is a standard loss for classification tasks. Minimization of the ENLL loss in Eq.~(\ref{eq:VI}) is usually archived via stochastic-gradient-based methods implemented on the Monte Carlo sampled model. 
\\

\textbf{Deep ensembles.} Conventional maximum likelihood training of a neural network can be naturally described within the Bayesian variational inference picture~\cite{Graves2011}. Deep ensembles can be explicitly reformulated within the Bayesian setting as well~\cite{Hoffmann2021} and are associated with ansatz 
\begin{align} 
\begin{split} 
&q_\omega(\theta) = \frac{1}{K}\sum^K_{k = 1} \, \mathcal{N}(\theta; \omega_k,  \sigma^2 I_{\text{dim} [\theta]}),\\
\end{split}
\label{eq:BDE}
\end{align}
where $K$ is the number of networks in the ensemble, $\omega$ consists of $K$ independent sets $\omega_k$ of parameters $\theta$, $k = 1, ..., K$. Conventional deep ensembles are associated with the case of infinitesimal (machine-precision) $\sigma$ so that the normal distributions in Eq.~(\ref{eq:BDE}) are sharply peaked and approach the delta function distributions $\delta(\theta - \omega_k)$.

The ENLL contribution to the cost function in Eq.~(\ref{eq:VI}) simplifies to a sum of independent loss terms associated with each ensemble member, weighted by factor $1/K$. In this case the KL divergence against the standard normally-distributed prior $p(\theta) = \mathcal{N}(\theta; 0, \, \lambda^{-1} I_{\text{dim}[\theta]})$ reduces to~\cite{Hoffmann2021}
\begin{align} 
\begin{split}
 &\text{KL}( \, q_\omega(\theta) \,||\,p(\theta) \, )  =  \, \frac{1}{2} \text{dim}[\theta]\left(\lambda \sigma^2-\log \, \sigma^2 - 1 - \log \lambda \right)\\
 & + \underbrace{\frac{1}{2K} \sum^K_{k = 1} \lambda ||\omega_k||^2}_\text{L2 regularization} - \underbrace{\ \ \log K \ \ }_\text{\shortstack{KL loss reduction \\ due to ensembling}} + \underbrace{\ \ \ \text{RF}_{\phantom{2}}  \ \ }_\text{repulsive force},\\
\end{split}
\label{eq:BDE_KL}
\end{align}
 where the repulsive-force correction $\text{RF}$ becomes negligibly small as $\sigma$ approaches zero and was dropped in Ref.~\cite{Hoffmann2021}. In our SM~\cite{OBHL_SM} we show that the repulsive-force correction can be approximated by the following upper bound
 \begin{align} 
\begin{split}
 &\text{RF} \leq \frac{1}{K}\sum^K_{k=1} \sum_{k^\prime \neq \, k} \exp[-||\omega_k - \omega_ {k^\prime}||^2/(8 \, \sigma^2)].\\
\end{split}
\label{eq:RF}
\end{align}

The L2 regularization term in Eq.~(\ref{eq:BDE_KL}) corresponds to our choice of a 
 Gaussian prior distribution for the model parameters, other terms come from the entropy contribution $H(q_\omega(\theta))$ to $\text{KL}( \, q_\omega(\theta) \,||\,p(\theta) \, )$ that is independent of the prior choice. The KL loss reduction due to ensembling is represented here by the $[-\log K]$ term that decreases with the  ensemble size. Interestingly, Eq.~(\ref{eq:BDE_KL}) also contains a repulsive force in weight space promoting the ensemble members to be diverse. Diversity among the ensemble members is believed to be a key element for reaching good performance~\cite{Lakshminarayanan2019, Angelo2021}.  

\section{Stochastic Ensembles}
\label{sec5}

\textbf{Posterior landscape}. To motivate our construction we start with an intuitive perspective. Regular deep ensembles are believed to be effective at approximating the posterior because of their ability to probe multiple modes of the posterior landscape~\cite{Lakshminarayanan2019, Hoffmann2021}. This is in contrast to less effective single network stochastic methods that only sample from a single mode. MultiSWAG~\cite{Wilson2020} improves deep ensembles by virtue of Stochastic Weight Averaging protocol that finds flatter parameter regions in the basins of attraction uncovered by deep ensembles. The flatter regions are believed to more accurately represent the posterior~\cite{Izmailov2018, Maddox2019, Wilson2020}. Note, however, that the basins of attraction sampled by MultiSWAG may not be optimal and there can exist wider basins that may be less probable for regular training protocols to end up in. Stochastic ensembles are built from trained networks regularized by stochastic methods such as dropout. They are also multimodal but they only find basins of attraction that are robust to stochastic perturbations and sampling from wide, robust parameter regions is essential for constructing an efficient posterior approximation, cf. the outlined arguments for MultiSWAG.

The length scale of parameter regions probed by a stochastic method is defined by its hyperparameters: dropout networks with larger drop rates, for example, will look for larger robust parameter regions. There will always be a trade off between the region size and its quality in terms of the robustness. To avoid the need to manually tune the hyper parameters one can aim towards finding a non-parametric stochastic method capable of probing parameter regions on different scales. This motivated us to introduce a non-parametric version of dropout described in the next section. We also note that different stochastic methods probe the parameter space in different ways. For example dropout probes random hyperplane projections in the parameter space. 
\\

\textbf{Bayesian description.} A well known example of stochastic regularization methods is Monte Carlo dropout \cite{Folgoc2021}. For concreteness let us consider a single fully-connected layer with conventional (Bernoulli) dropout applied after some activation. The layer's input-output transformation, say from the node layer $x_l$ to the node layer $x_{l+1}$, can be expressed as follows 
\begin{align} 
\begin{split} 
x_{l+1} = \sigma(W_{l+1, \, l} \, x_{l}  + B_{l+1}) \circ z_{l+1},\\
\end{split}
\label{eq:DO}
\end{align}
 where $W_{l+1, \, l}$ and $B_{l+1}$ are the layer's weights and biases respectively, $\sigma(x)$ is the activation, and $z_{l+1, n} \sim \mathcal{B}(p_{l+1, n})$ are Bernoulli-distributed variables, namely $z_{l+1, n} = 1$ with probability $p_{l+1, n}$ and $z_{l+1, n} = 0$ with probability $(1 - p_{l+1, n})$ with $n = 1, ..., N$. Here $N$ is the layer's output dimension and $\circ$ denotes the element-wise (Hadamard) product.

The dropout input-output transformation in Eq.~(\ref{eq:DO}) can be equivalently described as Monte Carlo sampling from a family of Bernoulli functions. Recently it was shown that networks regularized by Bernoulli methods can be viewed as variational inference~\cite{Gal2015, Gal2016, Gal2017}. We combine these networks into an ensemble and obtain a generalization of the deep ensemble variational ansatz in Eq.~(\ref{eq:BDE}) that we call stochastic ensemble. For the example of a fully-connected dropout layer in Eq.~(\ref{eq:DO}), the stochastic ensemble ansatz is given by
\begin{align} 
\begin{split} 
&q_\omega(\theta_l) = \frac{1}{K}\sum_{k = 1}^K \prod^{N}_{n = 1} \, \hat{q}_{\omega_{l, n, k}}(\theta_{l, n}),\\
\end{split}
\label{eq:BSDON}
\end{align}
with
\begin{align} 
\begin{split} 
\hat{q}_{\omega_{l, n, k}}(\theta_{l, n}) &=  p^{(1)}_{l+1, \,n}  \, \mathcal{N}(\, \theta_{l, n}; \omega^{(1)}_{\,l, n, k}\, , \,  \sigma^2 \, I_{\text{dim} [\theta_{l, n}]}\,)\\
&+ \,  p^{(2)}_{l+1, \,n}  \, \mathcal{N}(\, \theta_{l, n}; \omega^{(2)}_{\,l, n, k}\, ,  \, \sigma^2 \, I_{\text{dim} [\theta_{l, n}]}\,),\\
\end{split}
\label{eq:BSDON_b}
\end{align}
where $n$ labels the layer's output nodes $x_{l+1}$, $\theta_{l, n}$ consists of all weights and bias connecting the layer's input $x_l$ to the layer's output node $x_{l+1, n} \,$,  and $\omega^{(i)}_{l, n, k}$ for $i = 1, 2$ are independent realizations of the network's parameters $\theta_{l, n}$, $p^{(i)}_{l,n}$ the Bernoulli probabilities for the two realizations, and $k = 1, ..., K$.

As in the case of regular deep ensembles, the ENLL loss reduces to a sum of independent single network losses weighted by $1/K$. Each such loss term and its gradients can be handled using conventional stochastic methods such as Monte Carlo sampling~\cite{Gal2016}. The KL divergence against the normal prior reduces to (see our SM~\cite{OBHL_SM}):
\begin{align} 
\label{eq:BPDEDO_KL}
\begin{split}
 &\text{KL}( \, q_\omega(\theta) \,||\,p(\theta) \, )  =  \, \frac{1}{2} \text{dim}[\theta]\left(\lambda \sigma^2-\log \, \sigma^2 - 1 - \log \lambda \right)\\
 & + \frac{1}{2K} \sum^K_{k = 1} \sum^N_{n = 1} \sum^2_{i = 1} \lambda \, p^{(i)}_{l+1, \,n} ||\omega^{(i)}_{\,l, n, k}||^2 - \log K \\
 & + \underbrace{\sum^{N}_{n = 1} \sum^2_{i = 1} p^{(i)}_{l+1, \,n} \log p^{(i)}_{l+1, \,n}}_\text{\shortstack{KL loss reduction \\due to stochasticity}}+ \text{RF}_2. 
\end{split}
\end{align}
The difference to regular deep ensembles in Eq.~(\ref{eq:BDE_KL}) is one additional term that describes the KL loss reduction corresponding to using stochastic over non-stochastic deep networks~\cite{Gal2016}. The repulsive force $\text{RF}_2$ is of a more complex form than in the regular deep ensemble case, see our SM~\cite{OBHL_SM}  for the derivation of the corresponding upper bound. The bound contains an accumulation of the RF contributions coming from every possible stochastic parameter realizations, weighted by the \text{probabilities} of obtaining these configurations. Note that it approaches zero for infinitesimal (machine-precision) $\sigma$.
\\

\textbf{Stochastic ensemble realizations}. The ansatz in Eqs.~(\ref{eq:BSDON}) and (\ref{eq:BSDON_b}) is equivalent to the dropout ensemble by assuming the limit of small $\sigma$, and taking $\omega^{(2)}_{\,l, n, k}$ to be always zero. Other types of stochastic ensembles, such as DropConnect ensembles, as well as other varieties of network layers, beyond the fully-connected layers, can be formulated similarly to Eq.~(\ref{eq:BSDON}). In our experiments we study the dropout and DropConnect realizations of stochastic ensembles, and denote them by SE1 and SE2 respectively. We also look at a realization of Eqs.~(\ref{eq:BSDON}) and (\ref{eq:BSDON_b}) where two sets of trainable parameters are exchanged with equal probability $p^{(1)}_{l+1, \,n} = p^{(2)}_{l+1, \,n} = 1/2$. This provides an non-parametric method (in the sense of no tunable probabilities), referred to as SE3, that can be applied to any layer type with just minor changes, including the output layer.
\\

\textbf{Limitations of the KL loss}.  The KL loss in Eqs.~(\ref{eq:BDE_KL}) and (\ref{eq:BPDEDO_KL}) suggest that ensembles of stochastic networks are expected to more accurately approximate the posterior than regular deep ensembles, especially for large models because the loss reduction due to stochasticity in Eq.~(\ref{eq:BPDEDO_KL}) scales with $N$. The KL divergence, however, is not a flawless measure. It is known to ignore parameter regions where the distribution $q_\omega(\theta)$ is small, and therefore may not penalize the discrepancy with some important parts of the posterior. An interesting improvement to the KL loss was studied in Ref.~\cite{Gal2017_2}.

We also stress that the total KL loss that is minimized during the training is not directly correlated with the performance metrics such as accuracy and calibration on the validation set. Smaller KL loss is expected to lead to better quantitative estimates, however, for a given task the quality of these estimates may depend on the data and considered metrics. 

\section{Numerical Results}

\subsection{Toy model}

To quantitatively compare different ensemble methods we first evaluate them on a toy classification problem. The advantage of using a simple toy model is that the posterior can be straightforwardly sampled using HMC. We implement the NUTS extension \cite{NUTS} of HMC and directly evaluate performance of ensemble methods by comparing their outcomes with the corresponding HMC posterior. The HMC posterior is constructed by stacking 4 independent HMC chains, each of 2000 samples. The details on our HMC (NUTS) implementation are given in the SM~\cite{OBHL_SM}.

\begin{figure}
	\centering
	\includegraphics[width=235pt]{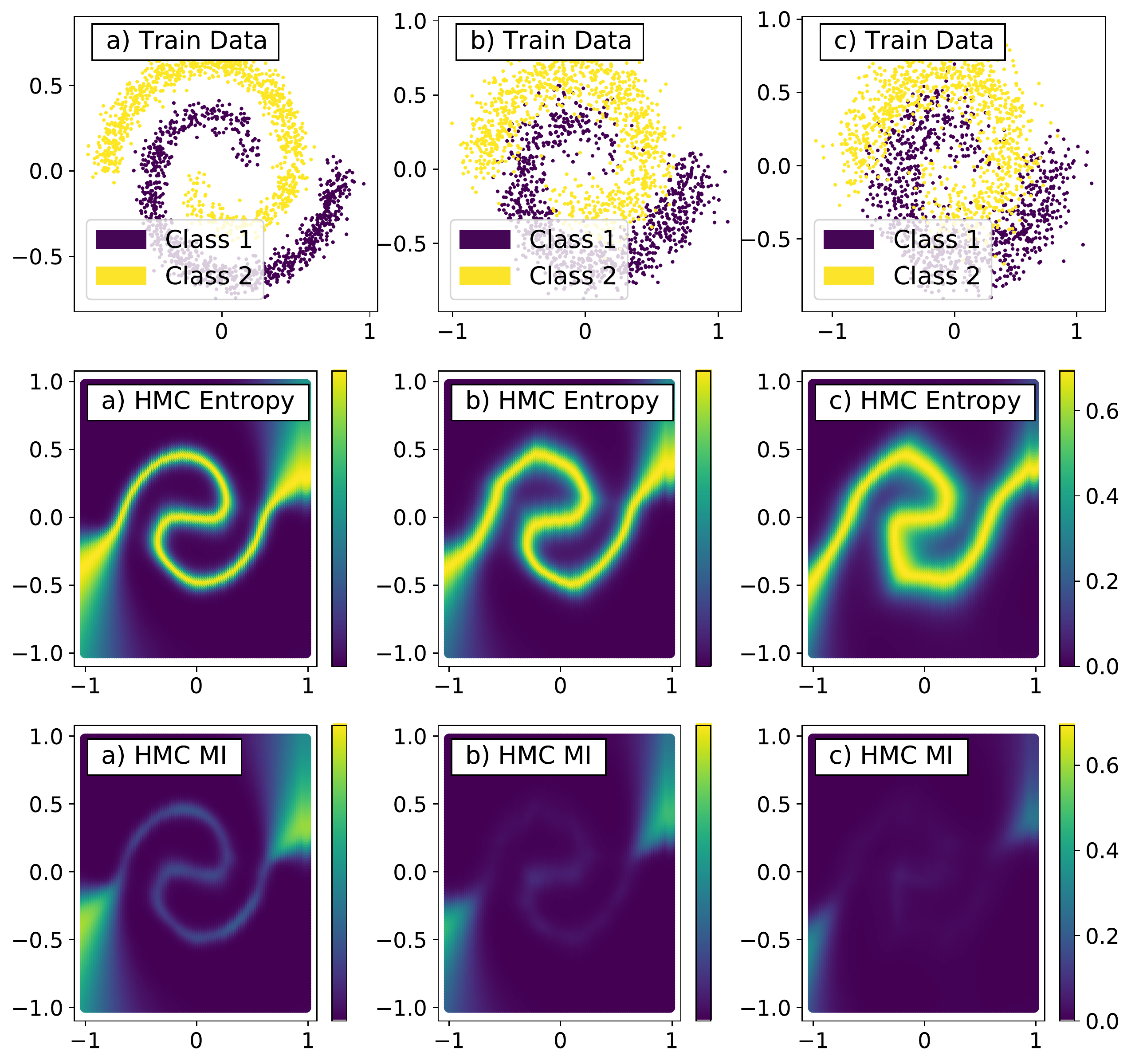}
	\caption{(a-c) Three different datasets of training data and corresponding to them HMC uncertainty estimates, entropy and mutual information. }
\label{fig:fig1}
\end{figure}

We consider three training datasets shown in Fig.~\ref{fig:fig1} \text{(a-c)}, where the 2D data is classified into two classes. The datasets represent different class separability levels, helping us to make a more comprehensive comparison of the ensemble methods. We employ a simple feed forward network with two hidden layers of 10 neurons, ReLU activations, and softmax output. The trained models are then tested on two uniformly sampled data domains, $\mathcal{D_\text{in}}=[-1, \, 1]^2$ (in-domain) and $\mathcal{D_\text{out}}=[-10, \, 10]^2$ (out-of-domain). We consider three types of stochastic ensembles in our experiments: Monte Carlo dropout (SE1), DropConnect (SE2), and non-parametric dropout (SE3).  More details on the model, training procedure and ensembles are provided in our SM~\cite{OBHL_SM}.
\\

\textbf{Uncertainty estimates.} In Fig.~\ref{fig:fig1} we also depict two important uncertainty estimates evaluated using the HMC posterior. The total predictive uncertainty can be estimated by computing the predictive posterior entropy that formally quantifies how much information on average is contained in the output~\cite{Shannon1948}. The second quantity is the (average) mutual Shannon information contained between network parameters and test data sample conditioned on the training dataset. It quantifies how the posterior distribution changes as we include a new data point to the training dataset~\cite{MacKay1992}. The mutual information describes only the epistemic part of the total uncertainty and it is an important quantity used in many deep learning applications, for example, in active learning\cite{reviewAL2021}. For more details on these two quantities, and our implementation, see the SM~\cite{OBHL_SM}.

The toy classification problem is a good illustration for the concepts of aleatoric (data) uncertainty and epistemic (model) uncertainty. For the boundary regions between the two classes, the prediction is expected to exhibit high uncertainty that is mostly of aleatoric origin. Our HMC results in Fig.~\ref{fig:fig1} are in agreement with the expectation: The total entropy correctly identifies regions of high uncertainty, both the boundary regions (high aleatoric uncertainty) and certain regions that were not used in the training (high epistemic uncertainty). The mutual information is seen to be large only in the regions of high epistemic uncertainty. The aleatoric (epistemic) uncertainty is seen to increase (decrease) with the class mixing level.
\\ 

\begin{table}
\setlength{\tabcolsep}{0.5em} 
\begin{tabular}{c ||c c c c} 
 \hline  
   & \shortstack{Entropy\\ {1e-3}}\rule{0pt}{4.5ex} & \shortstack{MI\\ {1e-3}}\rule{0pt}{4.5ex}  & \shortstack{Agr\\ {1e-2}}\rule{0pt}{4.5ex}  & \shortstack{Var\\ {1e-2}}\rule{0pt}{4.5ex}  \\
 \hline\hline
 & \multicolumn{4}{c}{ \rule{0pt}{3ex} Toy-a (in-domain / out-of-domain)} \\
 \hline
  \shortstack{Regular\\ \phantom{Regular} } \rule{0pt}{4.5ex} & \shortstack{0.51 \\ 1.18} & \shortstack{ 0.42\\1.16 } & \shortstack{ 98.3 \\ 95.2} & \shortstack{ 2.76 \\ 7.49} \\
   \hline
  \shortstack{MultiSWA\\\phantom{SWAG}} \rule{0pt}{4.5ex}& \shortstack{ 0.71 \\  1.32}& \shortstack{ 0.48\\ 1.31} & \shortstack{  97.30\\ 94.01}& \shortstack{ 4.00\\ 8.46}\\
  
 \hline
  \shortstack{NP Dropout\\(SE3)} \rule{0pt}{4.5ex}& \textbf{\shortstack{ 0.46\\ 1.14}}& \textbf{\shortstack{ 0.39\\ 1.13}} & \textbf{\shortstack{ 98.4 \\ 95.5}}& \textbf{\shortstack{ 2.59\\ 7.00}}\\
 \hline  \hline
 & \multicolumn{4}{c}{ \rule{0pt}{3ex} Toy-b (in-domain / out-of-domain)} \\
 \hline
  \shortstack{Regular\\ \phantom{Regular} } \rule{0pt}{4.5ex} & \shortstack{ 0.30\\ 0.89}  & \shortstack{0.19 \\ 0.87} & \shortstack{ \textbf{98.9}\\95.4 } & \shortstack{ 1.61\\ 5.44}\\
 \hline
 \shortstack{MultiSWA\\\phantom{SWAG}} \rule{0pt}{4.5ex}& \shortstack{ 0.49\\ 1.06}& \shortstack{ 0.21\\ 1.03} & \shortstack{  97.7\\ 94.4}& \shortstack{ 2.89\\ 6.74}\\
  
 \hline
 
  \shortstack{NP Dropout\\(SE3)} \rule{0pt}{4.5ex}& \textbf{\shortstack{ 0.25\\ 0.73}}  & \textbf{\shortstack{ 0.16\\ 0.71}}& \textbf{\shortstack{ 98.9\\ 95.8}} & \textbf{\shortstack{ 1.41\\ 4.51}}\\
 \hline  \hline
  & \multicolumn{4}{c}{ \rule{0pt}{3ex} Toy-c (in-domain / out-of-domain)} \\
 \hline
  \shortstack{Regular\\ \phantom{Regular} } \rule{0pt}{4.5ex} & \shortstack{ 0.23\\ 0.86} & \shortstack{ 0.12\\ 0.83} & \shortstack{ \textbf{99.1}\\ \textbf{96.8}} & \shortstack{ \textbf{1.19}\\ 4.39}\\
 \hline
 \shortstack{MultiSWA\\\phantom{SWAG}} \rule{0pt}{4.5ex}& \shortstack{ 0.32\\0.90 }& \shortstack{ 0.10\\0.86 } & \shortstack{  98.1\\ 95.5}& \shortstack{ 2.01\\5.01 }\\
  
 \hline
  \shortstack{NP Dropout\\(SE3)} \rule{0pt}{4.5ex}& \textbf{\shortstack{0.22 \\ 0.76}} & \textbf{\shortstack{ 0.11\\ 0.73}}& \shortstack{99.0 \\ 96.7}& \shortstack{ 1.20\\ \textbf{3.86}}\\
 \hline  
\end{tabular}
\caption{Quantitative comparison of predictions obtained from HMC, regular ensemble, MultiSWA, and non-parametric dropout ensemble SE3. The tests are done using data from $\mathcal{D_\text{in}}=[-1, \, 1]^2$ (in-domain, top rows) and $\mathcal{D_\text{out}}=[-10, \, 10]^2$ (out-of-domain, bottom rows). We considered all three different toy datasets from Fig.~\ref{fig:fig1}. The considered metrics are agreement, variance, mean absolute difference of entropy and mutual information estimates computed in respect to the HMC runs. All variances are orders of magnitude smaller than quoted results given the large ensembles.
}
\label{Tab:toy}
\end{table}

\textbf{Performance of the ensemble methods.} We quantitatively evaluate performance of the ensemble methods by looking at the agreement and variance between the ensemble posteriors and  “ground truth” HMC posterior. For an explicit definition of these two quantities see our SM~\cite{OBHL_SM} or Ref.~\cite{Izmailov2021}. We also report baseline results for the regular (non-stochastic) deep ensembles and MultiSWA ensemble method~\cite{Wilson2020}. To quantify the quality of uncertainty estimates we also look at the mean absolute differences computed between the HMC and ensemble estimates. All three classification datasets depicted in Fig.~\ref{fig:fig1} (a-c) are considered. For each dataset, we compute metrics on data from $\mathcal{D}_\text{in}$ (in-domain) and $\mathcal{D}_\text{out}$ (out-of-domain). The size of each ensemble is 1024. Here we always take only one Monte Carlo inference per trained ensemble member, in this way forcing all types of ensembles to use the same resources at test time. In accordance with this constraint, we implement the test-time efficient MultiSWA instead of MultiSWAG~\cite{Wilson2020}, see our SM~\cite{OBHL_SM} for implementation details.

The dropout (SE1) and DropConnect (SE2) ensembles
were found to not perform as good as the regular and non-parametric dropout (SE3) ensembles, highlighting importance of correctly picking the stochastic type for a particular task. For our toy problems we could anticipate this behaviour by
looking at the training KL loss exhibiting high fluctuations for SE1 and SE2. The KL loss of SE3 and non-stochastic networks did not fluctuate as much and resulted in smaller training KL loss values.  

We present results corresponding to the non-parametric dropout (SE3), MultiSWA and regular ensembles in  Table~\ref{Tab:toy}. We observe that SE3 outperforms the other two methods, in most places with large margin. The average improvement with respect to the regular ensemble is approximately 10 percents in each of the metrics. Interestingly, the improvement differs between the different versions of the classification dataset and drops with the mixing between the training datasets. We attribute this to the difficulty of stochastic methods to fit the boundary between mixed classes. We also note that for the case with the largest mixing, SE3 and regular ensembles are indistinguishable in agreement and variance for in-domain data. Surprisingly, MultiSWA does not perform better than the regular ensemble baseline.

\subsection{CIFAR}

Let us consider now more realistic data and focus on the classification of images. The CIFAR dataset \cite{Krizhevsky2009} consists of 60000 low-resolution (32 by 32) RGB images classified into 10 (CIFAR-10) or 100 (CIFAR-100) classes.  Complexity of this image dataset makes the full-batch HMC computations highly nontrivial and computationally demanding. Only recently the full-batch HMC method has been evaluated on CIFAR for large residual networks~\cite{Izmailov2021}. In this section we train the much less demanding stochastic and non-stochastic ensembles on the CIFAR dataset and compare our results to the HMC data from Ref.~\cite{Izmailov2021}. We employ the same network architecture: A ResNet-20-FRN residual network of depth 20 with batch normalization layers replaced with filter response normalization (FRN). The HMC posteriors are directly loaded from the publicly available resource~\cite{Izmailov2021}.

Bayesian variational inference fixes some aspects of the training procedure such as weight decay and dispense with the need for early stopping. Our training procedure is therefore different from the implementation in Ref.~\cite{Izmailov2021} leading to different results for regular ensembles. Three types of stochastic ensembles are used in our CIFAR tests: Monte Carlo dropout (SE1), DropConnect (SE2), and non-parametric dropout (SE3). The drop rates of SE1 and SE2 were tuned to target the HMC results, rather than targeting best performance in terms of accuracy and loss. Each ensemble consists of 50 networks and we use strictly one Monte Carlo inference per ensemble member at test time, i.e. in total 50 inferences per ensemble. This restricts the methods to use the same resources. We also considered increasing the number of inferences per member, obtaining only a minor change in the performance. We report these results in our SM~\cite{OBHL_SM}. We also trained multiple realizations of the MultiSWA protocol corresponding to different hyperparameter sets. In the following we only present the closest MultiSWA realization to HMC. For the details on our hyperparameter selection see the SM~\cite{OBHL_SM}.
\\

\begin{table}
\setlength{\tabcolsep}{0.5em} 
\begin{tabular}{c ||c c c c c c } 
 \hline  
   & \shortstack{Acc\\\phantom{1}} \rule{0pt}{2.5ex}  & \shortstack{Loss\\ \phantom{1}} & \shortstack{ECE\\ \phantom{1}} & \shortstack{Agr\\{1e-2}}\rule{0pt}{4.5ex}  & \shortstack{Var\\{1e-2}}\rule{0pt}{4.5ex}  & \shortstack{ODD\\ \phantom{1}}  \\
 \hline\hline
 & \multicolumn{6}{c}{ \rule{0pt}{3ex} CIFAR-10 } \\
 \hline
 \shortstack{HMC\\3-chains}  \rule{0pt}{4.5ex}& \shortstack{ 90.68\\ \phantom{90.7}} & \shortstack{  0.307 \\ \phantom{90.7}}& \shortstack{0.059 \\ \phantom{90.7}}& \shortstack{100 \\ \phantom{90.7}} & \shortstack{0.0 \\ \phantom{90.7}}& \shortstack{85.3 \\ \phantom{90.7}} \\ 
 \hline

   \shortstack{Regular\\\phantom{(1-chain)}}  \rule{0pt}{4.5ex}& \shortstack{ 88.82\\ {\footnotesize $\pm$ 00.03}} & \shortstack{0.339\\ {\footnotesize $\pm$ 0.001}} & \shortstack{0.028\\ {\footnotesize $\pm$ 0.001}} & \shortstack{92.5\\ {\footnotesize $\pm$ 0.1}} & \shortstack{10.0\\ {\footnotesize $\pm$ 0.2}} & \shortstack{83.7\\ {\footnotesize $\pm$0.2}}\\
 \hline
\shortstack{Multi\\SWA}  \rule{0pt}{4.5ex}& \shortstack{88.49\\ {\footnotesize $\pm$ 0.10}} & \shortstack{0.372\\ {\footnotesize $\pm$ 0.024}}  & \shortstack{0.051 \\ {\footnotesize $\pm$ 0.025}} & \shortstack{92.3\\ {\footnotesize $\pm$ 0.1}}&  \shortstack{11.6\\ {\footnotesize $\pm$ 1.4}}& \shortstack{83.0\\ {\footnotesize $\pm$1.4}} \\
 \hline
   \shortstack{Dropout\\ (SE1)}  \rule{0pt}{4.5ex}& \shortstack{ \textbf{90.82}\\ \textbf{\footnotesize $\pm$ 0.03}} & \shortstack{\textbf{0.301}\\ \textbf{\footnotesize $\pm$ 0.001}}  & \shortstack{\textbf{0.057}\\ \textbf{\footnotesize $\pm$ 0.001}} & \shortstack{\textbf{94.1}\\ \textbf{\footnotesize $\pm$ 0.1}} & \shortstack{\textbf{8.0}\\ \textbf{\footnotesize $\pm$ 0.1}} &  \shortstack{\textbf{85.5}\\ \textbf{\footnotesize $\pm$0.1}} \\
  \hline
   \hline
 & \multicolumn{6}{c}{ \rule{0pt}{3ex} CIFAR-100 } \\
 \hline

  \shortstack{HMC\\3-chain}  \rule{0pt}{4.5ex}& \shortstack{ 67.39\\\phantom{1}} & \shortstack{1.205\\\phantom{1}} &  \shortstack{0.131\\\phantom{1}} & \shortstack{100\\ \phantom{1}}& \shortstack{0.0\\ \phantom{1}}&  \shortstack{72.5\\\phantom{1}}\\
 \hline
   \shortstack{Regular\\\phantom{(1-chain)}}  \rule{0pt}{4.5ex}& \shortstack{ 62.84\\ {\footnotesize $\pm$ 0.16}} & \shortstack{1.477\\ {\footnotesize $\pm$ 0.001}}  & \shortstack{ 0.157\\ {\footnotesize $\pm$ 0.002}} & \shortstack{71.4\\ {\footnotesize $\pm$ 0.1}}& \shortstack{29.8\\ {\footnotesize $\pm$ 0.1}}& \shortstack{71.5\\ {\footnotesize $\pm$0.1}} \\
\hline

\shortstack{Multi\\SWA}  \rule{0pt}{4.5ex}& \shortstack{61.86\\ {\footnotesize $\pm$ 0.6}} & \shortstack{ 1.526\\ {\footnotesize $\pm$ 0.018}}  & \shortstack{0.164\\ {\footnotesize $\pm$ 0.002}} & \shortstack{70.3\\ {\footnotesize $\pm$ 0.1}}&  \shortstack{31.3\\ {\footnotesize $\pm$ 0.1 }}& \shortstack{71.6\\{\footnotesize $\pm$0.1 }}\\

 \hline
   \shortstack{Dropout\\ (SE1)}  \rule{0pt}{4.5ex}& \shortstack{\textbf{68.49}\\ \textbf{\footnotesize $\pm$ 0.18}} & \shortstack{\textbf{1.167}\\ \textbf{\footnotesize $\pm$ 0.001}} & \shortstack{\textbf{0.124}\\ \textbf{\footnotesize $\pm$ 0.002}}& \shortstack{\textbf{77.2}\\ \textbf{\footnotesize $\pm$ 0.1}}&  \shortstack{\textbf{22.2}\\ \textbf{\footnotesize $\pm$ 0.1}}&  \shortstack{\textbf{73.5}\\ \textbf{\footnotesize $\pm$ 0.1}} \\
  \hline
\end{tabular}
\caption{Prediction accuracy (acc), test log-likelihood loss (loss), expected calibration error (ECE), agreement (agr), variance (var) and out-of-domain detection (ODD) for different inference methods trained on the CIFAR datasets. The HMC results are computed using the published HMC chains from Ref.~\cite{Izmailov2021}. Variances are calculated over independent runs.}
\label{Tab:CIFAR_NDA}
\end{table}

\textbf{Stochastic ensembles.} We found our Monte Carlo dropout ensembles (SE1) to produce more accurate posteriors than DropConnect (SE2) and non-parametric dropout (SE3) ensembles. We therefore present only results corresponding to SE1. We could anticipate SE1 performing better than SE2 and SE3 from the training KL loss analysis. 
\\

\textbf{Test performance metrics.} In Table~\ref{Tab:CIFAR_NDA} we present the test accuracy, test loss, expected calibration error (ECE), agreement, variance, and out-of-distribution detection (ODD) evaluated for different types of ensembles. For metric definitions see our SM~\cite{OBHL_SM}. From Table~\ref{Tab:CIFAR_NDA} we read off that the Monte Carlo dropout ensembles (SE1) provide more accurate posteriors than the baseline MultiSWA and regular deep ensembles. 
\\

\begin{figure}
	\centering
	\includegraphics[width=230pt]{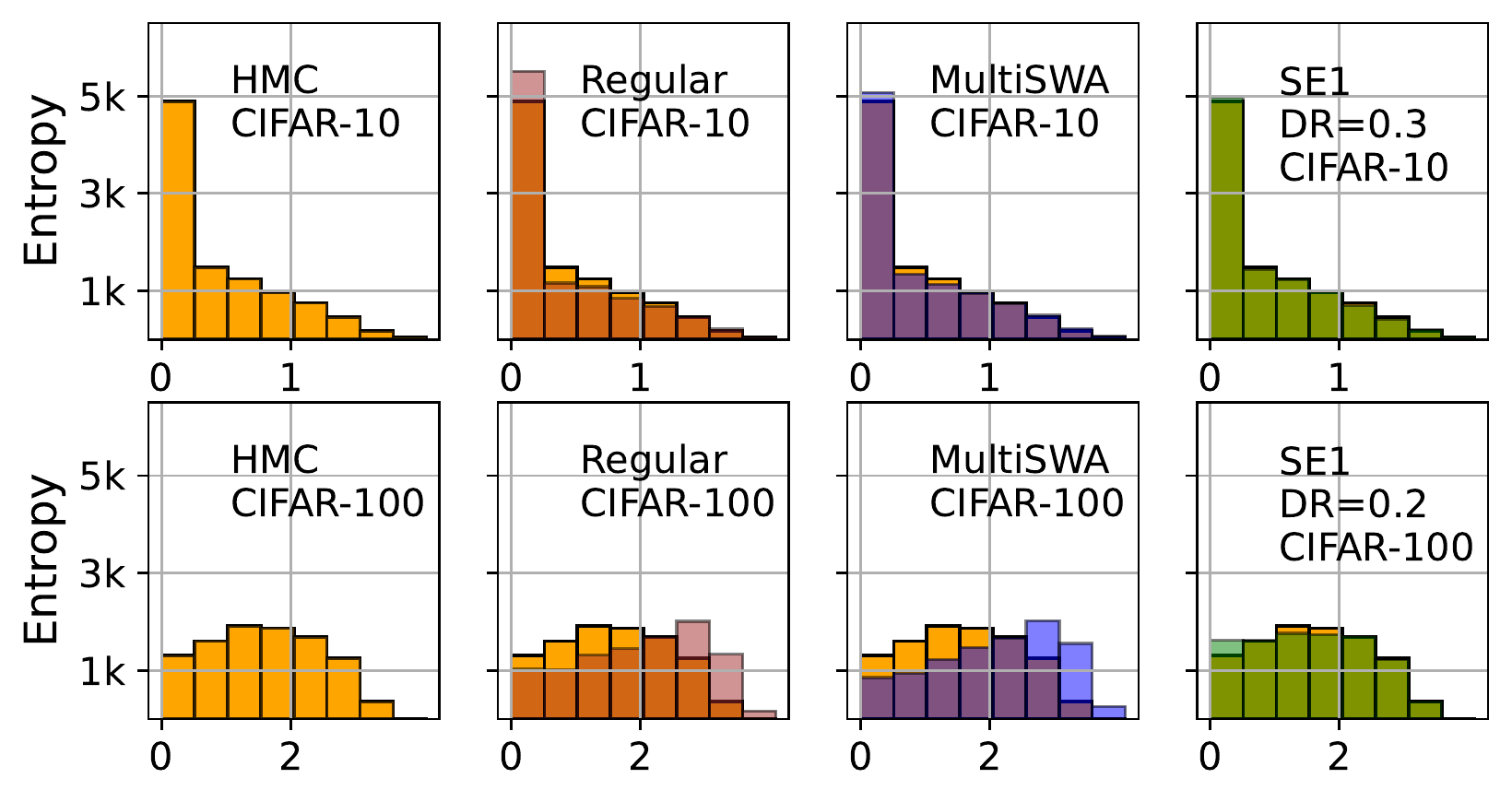}
	\caption{Histogram plots of predictive uncertainty computed using different ensemble methods. The HMC baselines are shown in orange in every subplot for better visualisation contrast. }
\label{fig:fig3}
\end{figure}

\textbf{Predictive entropy.} In Fig.~\ref{fig:fig3} we present the predictive entropy computed for different ensembles trained on CIFAR. SE1 is seen to match more accurately the HMC distribution of entropy values than regular deep ensembles. We also note that MultiSWA and SE1 perform equally well on CIFAR-10 but SE1 is more accurate on CIFAR-100.   
\\

\begin{figure}
	\centering
	\includegraphics[width=230pt]{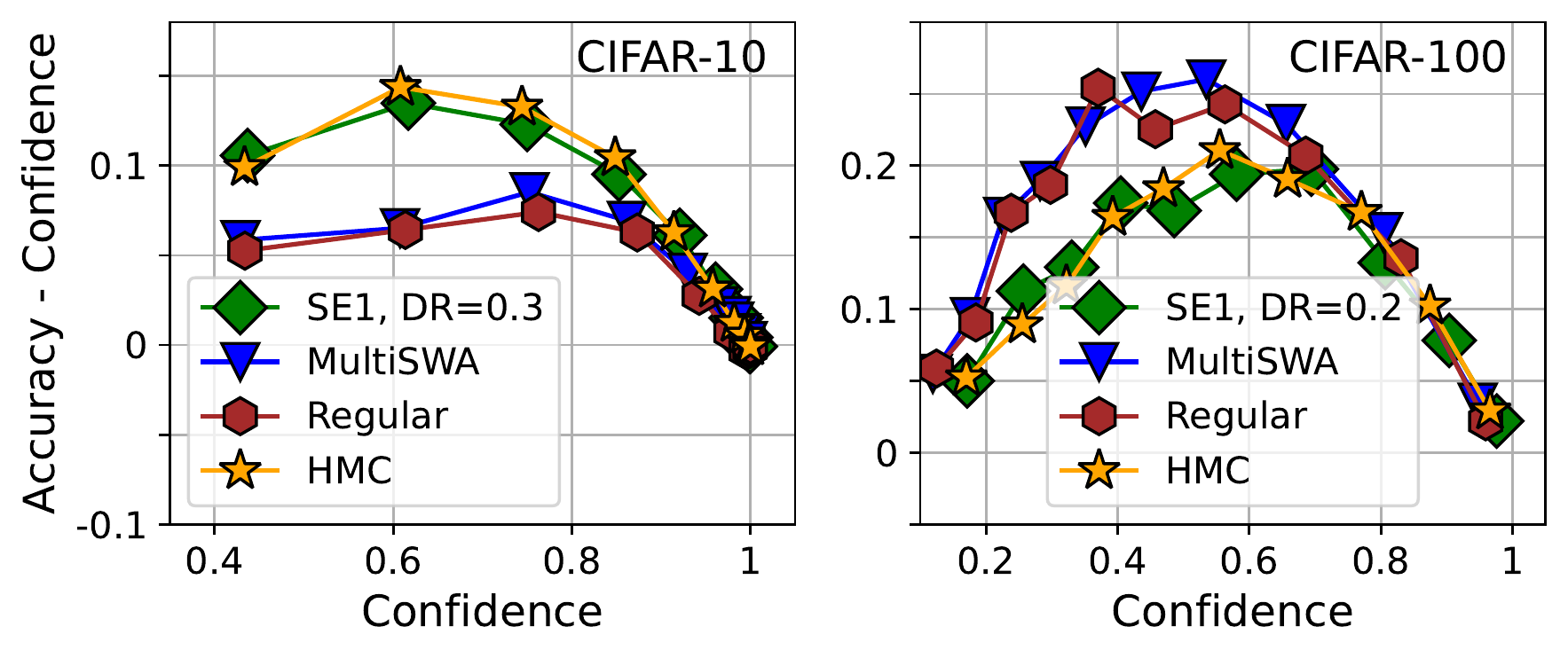}
	\caption{Calibration curves associated with different ensembles trained and evaluated on CIFAR.}
\label{fig:fig4}
\end{figure}

\textbf{Calibration curve.} An accurate approximation of Bayesian posterior is expected to be equally calibrated as HMC. In Fig.~\ref{fig:fig4} we depict calibration curves associated with different ensemble methods. All methods are found to be relatively poorly calibrated. The SE1 ensembles are seen to be the closest to HMC.
\\

\begin{figure}
	\centering
	\includegraphics[width=230pt]{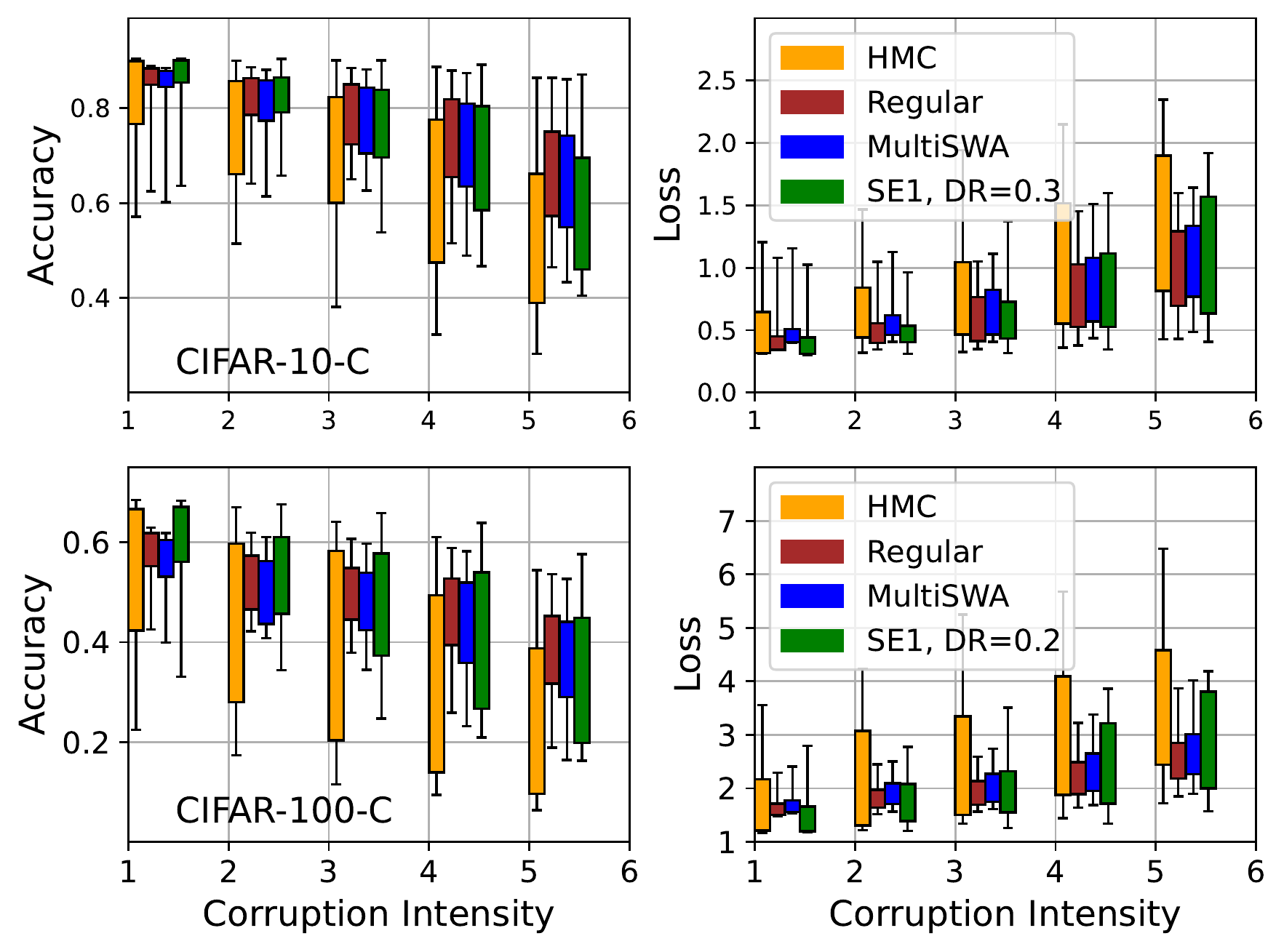}
	\caption{Evaluation on CIFAR-C. The CIFAR test sets are corrupted in 16 different ways at various intensities on the scale of 1 to 5. The error bars depict min and max of the corresponding metrics over each corruption, with the boxes indicating quartiles.}
\label{fig:fig5}
\end{figure}

\textbf{Robustness to distribution shift.} To test robustness to distribution shifts we evaluate the performance of different methods on the CIFAR-C and CIFAR-100-C datasets~\cite{Hendrycks2019}. These datasets contain CIFAR images altered using different types of corruptions with varying intensities. We use the same 16 corruptions as in Ref.~\cite{Izmailov2021}. For each method and corruption we compute the accuracy, loss, ECE, agreement, and variance, and then average over all the corrupted datasets. The corresponding data is presented in Table~\ref{Tab:CIFAR_NDA_C}, where SE1 is seen to again be closer to HMC in terms of the agreement and variance. In Fig.~\ref{fig:fig5} we plot the accuracy and loss resolved over different corruption types and intensities. SE1 is found to be more sensitive to the distribution shifts than the regular ensembles and MultiSWA but still not as much as HMC~\cite{Izmailov2021}. 
\\

\begin{table}
\setlength{\tabcolsep}{0.5em} 
\begin{tabular}{c || c c c c c} 
 \hline  
   & \shortstack{Acc\\ \phantom{[1e-2]}} \rule{0pt}{2.5ex}  & \shortstack{Loss\\ \phantom{1e-2}} & \shortstack{ECE\\ \phantom{1e-2}} & \shortstack{Agr\\ {1e-2}}\rule{0pt}{4.5ex}  & \shortstack{Var\\ {1e-2}}\rule{0pt}{4.5ex}  \\
 \hline\hline
 & \multicolumn{5}{c}{ \rule{0pt}{3ex} CIFAR-10-C (mean over corruptions)} \\
 \hline
 \shortstack{HMC\\3-chains}  \rule{0pt}{4.5ex}& \shortstack{ 71.05\\ \phantom{71.04}} & \shortstack{  0.879 \\ \phantom{90.7}}& \shortstack{0.079 \\ \phantom{90.7}}& \shortstack{100 \\ \phantom{90.7}} & \shortstack{0.0 \\ \phantom{90.7}}  \\ 
 \hline

   \shortstack{Regular\\\phantom{(1-chain)}}  \rule{0pt}{4.5ex}& \shortstack{ 78.48\\ {\footnotesize $\pm$ 0.21}} & \shortstack{\, 0.647\\ {\footnotesize $\pm$ 0.006}} & \shortstack{\, 0.042\\ {\footnotesize $\pm$ 0.001}} & \shortstack{79.8\\ {\footnotesize $\pm$ 0.3}} & \shortstack{20.6\\ {\footnotesize $\pm$ 0.4}}  \\
 \hline
\shortstack{Multi\\SWA}  \rule{0pt}{4.5ex}& \shortstack{ 77.43\\ {\footnotesize $\pm$ 0.24}} & \shortstack{\, 0.690\\ {\footnotesize $\pm$ 0.019}} & \shortstack{\, 0.058\\ {\footnotesize $\pm$ 0.013}} & \shortstack{80.2\\ {\footnotesize $\pm$ 0.2}} & \shortstack{20.9\\ {\footnotesize $\pm$ 0.2}}  \\
 \hline
   \shortstack{Dropout\\ (SE1)}  \rule{0pt}{4.5ex}& \shortstack{\textbf{76.32}\\ \textbf{\footnotesize $\pm$ 0.29}} & \shortstack{\textbf{0.713}\\ \textbf{\footnotesize $\pm$ 0.002}} & \shortstack{\textbf{0.064}\\ \textbf{\footnotesize $\pm$ 0.001}} &  \shortstack{\textbf{83.1}\\ \textbf{\footnotesize $\pm$ 0.1}} & \shortstack{\textbf{16.4}\\ \textbf{\footnotesize $\pm$ 0.1}} \\ 
  \hline
   \hline
 & \multicolumn{5}{c}{ \rule{0pt}{3ex} CIFAR-100-C (mean over corruptions)} \\
 \hline

  \shortstack{HMC\\3-chain}  \rule{0pt}{4.5ex}& \shortstack{ 42.10\\\phantom{1}} & \shortstack{2.610\\\phantom{1}} &  \shortstack{0.113\\\phantom{1}} & \shortstack{100\\ \phantom{1}}& \shortstack{0.0\\ \phantom{1}}\\
 \hline
   \shortstack{Regular\\\phantom{(1-chain)}}  \rule{0pt}{4.5ex}& \shortstack{49.33\\ {\footnotesize $\pm$ 0.13}} & \shortstack{\, 2.067\\ {\footnotesize $\pm$ 0.009}} & \shortstack{\, 0.115\\ {\footnotesize $\pm$ 0.001}} & \shortstack{53.3\\ {\footnotesize $\pm$ 0.3}} & \shortstack{39.7\\ {\footnotesize $\pm$ 0.2}}  \\

 \hline

\shortstack{Multi\\SWA}  \rule{0pt}{4.5ex}& \shortstack{\textbf{47.48}\\ \textbf{\footnotesize $\pm$ 0.18}} & \textbf{ \shortstack{\, 2.153\\ {\footnotesize $\pm$ 0.011}}} & \textbf{\shortstack{\, 0.112\\ {\footnotesize $\pm$ 0.001}}} & \shortstack{53.4\\ {\footnotesize $\pm$ 0.2}} & \shortstack{39.9\\ {\footnotesize $\pm$ 0.2}}  \\

 \hline
   \shortstack{Dropout\\ (SE1)}  \rule{0pt}{4.5ex}& \shortstack{48.35\\ \footnotesize $\pm$ 0.05} & \shortstack{{2.115}\\ {\footnotesize $\pm$ 0.003}} & \shortstack{{0.089}\\ {\footnotesize $\pm$ 0.001}}& \shortstack{\textbf{61.1}\\ \textbf{\footnotesize $\pm$ 0.2}}&  \shortstack{\textbf{32.0}\\ \textbf{\footnotesize $\pm$ 0.1}}\\
  \hline
\end{tabular}
\caption{Performance rates corresponding to different ensemble methods evaluated on CIFAR-C. The CIFAR test datasets are corrupted in 16 different ways at various intensities on the scale of 1 to 5. We average the results over all $16 \cdot 5$ corrupted datasets and report variances over independent runs. The HMC data was computed using the published HMC chains from Ref.~\cite{Izmailov2021}. }
\label{Tab:CIFAR_NDA_C}
\end{table}

{\textbf{Comparison to stochastic gradient Monte Carlo (SGMCMC) approximate posteriors.}}
Our SE1 is closer to HMC than published results \cite{Izmailov2021} for SGMCMC methods in terms of accuracy, loss, ECE, agreement, variance. SE1 also provides closer posteriors to HMC for CIFAR-10-C evaluations, while also being more computationally efficient.

\section{Conclusions}
    We introduced a variational ansatz for stochastic ensembles of neural networks. Common Bayesian posterior approximations such as regular deep ensembles, Monte Carlo dropout and DropConnect can all be formulated as special cases of these distributional families. Our theoretical and numerical results suggest that the ensembles of stochastic neural networks can provide more accurate approximations to the Bayesian posteriors than other baseline methods.

    Using our variational ansatz we formulated a new type (SE3) of non-parametric stochastic method that can be applied to any layer type with only minor changes to incorporate parameter sharing between ensemble members.  For a simple toy model, we showed with HMC sampling of the Bayesian posterior that our SE3 ensembles provide closer posteriors to HMC than all other considered ensemble methods.

We evaluated accuracy, loss, ECE, ODD, entropy, agreement, variance and robustness to distribution shift for a ResNet-20-FRN architecture using different types of  ensembles in image classification tasks on CIFAR. We found that our dropout stochastic ensembles (SE1) are closer to the HMC posteriors than all other methods considered.

 In each of our numeric tests, all the stochastic methods were manually tuned except non-parametric dropout. To understand better what hyperparameters and stochastic method to use for achieving the most accurate posterior approximations would be highly beneficial.

{\bf Acknowledgements.} OB was supported by the Knut and Alice Wallenberg Foundation and the Swedish Research Council (grant 2017-05162). HL and BM were supported by Vetenskapsr\aa det (grants 2017-3865 and 2021-4452).

{\small
\bibliographystyle{ieee_fullname}
\bibliography{egbib}
}

\end{document}